# Generative Adversarial Networks for geometric surfaces prediction in injection molding

Performance analysis with Discrete Modal Decomposition

Pierre Nagorny, Thomas Lacombe, Hugues Favrelière, Maurice Pillet, Eric Pairel
SYMME laboratory
Savoie Mont Blanc University
Annecy, France
pierre.nagorny@univ-smb.fr, thomas.lacombe1@univ-smb.fr, hugues.favreliere@univ-smb.fr

Ronan Le Goff, Marlène Wali, Jérôme Loureaux
IPC Technical Center - Centre Technique Industriel de la Plasturgie et des Composites
Bellignat, France
ronan.legoff@ct-ipc.com

Patrice Kiener
InModelia
Paris, France
patrice.kiener@inmodelia.com

**Geometrical and appearance quality requirements set the limits of the current industrial performance in injection molding. To guarantee the product's quality, it is necessary to adjust the process settings in a closed loop. Those adjustments cannot rely on the final quality because a part takes days to be geometrically stable. Thus, the final part geometry must be predicted from measurements on hot parts. In this paper, we use recent success of Generative Adversarial Networks (GAN) with the pix2pix network architecture to predict the final part geometry, using only hot parts thermographic images, measured right after production. Our dataset is really small, and the GAN learns to translate thermography to geometry. We firstly study prediction performances using different image similarity comparison algorithms. Moreover, we introduce the innovative use of Discrete Modal Decomposition (DMD) to analyze network predictions. The DMD is a geometrical parameterization technique using a modal space projection to geometrically describe surfaces. We study GAN performances to retrieve geometrical parameterization of surfaces.**

*Keywords— Injection molding; Quality prediction; Thermography; Generative Adversarial Networks ; Discrete Modal Decomposition.*

## I. Introduction

Thermoplastics injection molding can produce complex parts in large quantities. Quality requirements are increasing to guaranty the final product functionality. The final part quality depends on multiple settings and external non-controllable factors, from raw material hygrometry to in-mold pressure. Thus, closed-loop process settings adjustments are needed to achieve optimal quality [1]. Moreover, the final part quality can only be measured on stabilized parts, days after the production. Molded parts need to cool down with internal mechanical constraints relaxation. Thus, it is not possible to measure part quality just after production. To achieve next part process settings adjustment in closed loop, hot parts must be measured, and final part quality must be inferred. It is interesting to measure the hot part and infer the final part quality. Measurements and settings adjustments computation must be done in the industrial cycle time (less than thirty seconds). Thermographic imagery is a fast and easy to set-up measurement, which can be used on the production line. Inference of specific final parts quality based on hot parts measurements can be achieve using regressive models or neural networks. A previously learned model is fast to execute for inference on a recent computer. Convolution Neural Networks [2] have been used to predict a continuous geometry measurement of final part from thermographic images of hot parts [3]. However, a unique geometrical value is not meaningful for a human operator, and even more information could be extract from thermographic images. Recent research about Generative Adversarial Networks [4] used as images translators are applicable on industrial problems. In this paper, we propose the use of a GAN to translate thermographic images into final surface geometry (Fig. 1).

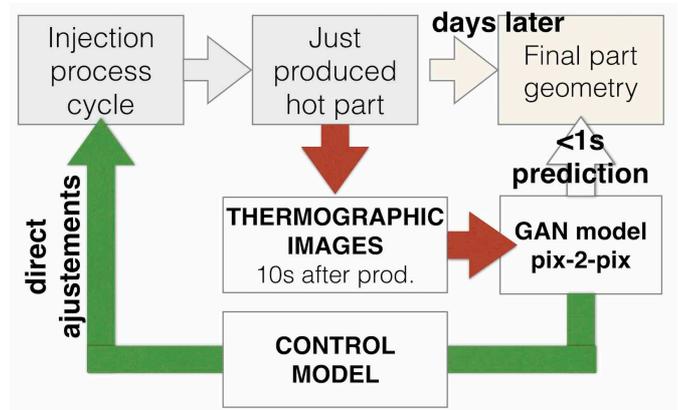

Fig. 1. Process ajustement based on thermography and GAN.

We measure the prediction performance with different images similarity comparison methods. Furthermore, we introduce the Discrete Modal Decomposition analysis [19] as an innovative tool to characterize the GAN generation performance to retrieve geometric parameterization. With some enhancement, this method could be proposed as a real-time tool to inspect the future part geometry, even in an augmented reality real-time overlay for human operators.

## II. RELATED WORK

Regulation of injection molding is a productive literature subject, using multiple advanced methods. Many works have been done using in-mold measurements with good success [5], but far lesser proposed to measure hot parts just after production. The serious disadvantage of in-mold measurements lays on its need of expensive sensors placed inside the mold with a custom mold design for wire routing. Besides, robotic is now heavily deployed in the industry. Thus, hot parts measurements can be done in-situ, in the production line and cycle time. If the measurement process is fast enough, every part can be measured. A robotic arm can move parts on different measurements stands. Thermography imagery, 3D scanning, weighing or photography are easy to set up. Then, mathematical models must be construct between hot parts measurements and final parts quality. Neural networks models can benefit from multi modal sensors fusion [6]. In this study, we work specifically on thermographic images as model input. Generative Adversarial Networks were introduced in 2014 and have since shown success in very specific image processing applications. Recent research used GAN to translate images, after the learning of a transformation model on pair images (pix2pix), or even with no pair prior at all (Cycle GAN) [7].

## III. METHODS

### A. Experimental dataset creation

Under industrial setting, it is usually difficult to create a big dataset. This paper aims to use a GAN network architecture to solve this limit. Generative networks are known to require a small dataset. We actually have a dataset with ten times fewer elements than classical dataset used for GAN. Our study is exploratory and aims to study the performance of GAN in industrial constraints.

The dataset creation was done using a Taguchi orthogonal L12 Design of Experiments [8], with twelve different trials. Furthermore, we choose to put two parts of each trial in the training dataset. An exception was done for on trial which parts are not at all in the training dataset but only in the validation dataset. These special not learned parts will be used for further validation of the network performance.

We use a robotic arm to grab the part right after molding and to show them in front of a thermographic camera. A still frame was acquired for each part 10 seconds after the ejection of the part Then, we have been waiting a month before geometrical measurements to be sure that parts were geometrically stable. Surfaces were scanned on a confocal optical profilometer (Altimet AltiSurf 520). The scanning takes one hour; this duration is the actual limit to the size of our dataset.

From the nearly two hundred parts we have produced and measured on-line, we have actually only scanned 37 parts: 23 parts are used as a training dataset and 14 parts are used for validation.

### B. Dataset preprocessing

The robotic arm and the vacuum prehension device were not enough repeatable to guarantee pixel to pixel position similarity between measurements. Thus, thermographic images were stabilized using the Lucas-Kanade optical flow tracking algorithm [9]. Images were normalized on the minimum and the maximum values of all images. Images were then cropped and scaled to 71 by 71 pixels in grayscale, without multiple color channels. Finally, images are upsampled (bicubic interpolation) to 128x128 to be used as networks input. Upsampling the small definition acts as a 1.8x1.8 pixels Gaussian filter. We voluntary choose small size images to simulate industrial constraint where multiple parts must be imaged with a unique camera, leading to a small image definition peer part. After normalization and reduction to 8 bits images the thermographic image has a resolution of 0.902 degrees Celsius. After normalization and reduction to 8 bits images, the height map images have a resolution of 1.57 μm. This is quite low, but it is acceptable to characterize low frequency geometric form defects. In our case, better results in term of geometry precision could be obtained with floating points input instead of 8 bits images, but the network size and training time should be far bigger.

### C. Pix2pix Generative Adversarial Networks architecture

A "Generative Adversarial Network" is composed by a generator and a discriminator which are both trained sequentially one after the other. The generator is trained to produce realistic images from a random normal distribution. Then, the discriminator must distinguish the "fake" generated images from the real images. At each training loop, the generator gets the difference of the output between the discriminator and the "fake" generated sample. The generator is train to increase the error of the discriminator by producing perfect counterfeit image. The discriminator is train on real images. The convergence of the network is achieved when the generator and the discriminator reach an equilibrium point between counterfeit performances and detection.

The pix2pix framework [7] use a deep convolutional GANs (DCGANs) architecture proposed by Radford, Metz and Chintala [10]. Pix2pix also proposes the use of conditional GAN [11], which learn their loss function from the observation of the inputted images. This framework simplifies many previous works and propose to use a "U-Net" architecture [12]. The U-Net architecture has skip connection between each center symmetric layers. Each skip connection concatenates channels from layer $i$ with channels from layer $n - i$, with n the total number of layers in the network. Pix2pix also use a convolutional classifier, which work on small patches of the input image (PatchGAN). Thus, the convolutional layer will be only receptive to pattern at the patch size scale. The patch size choice must be investigated.

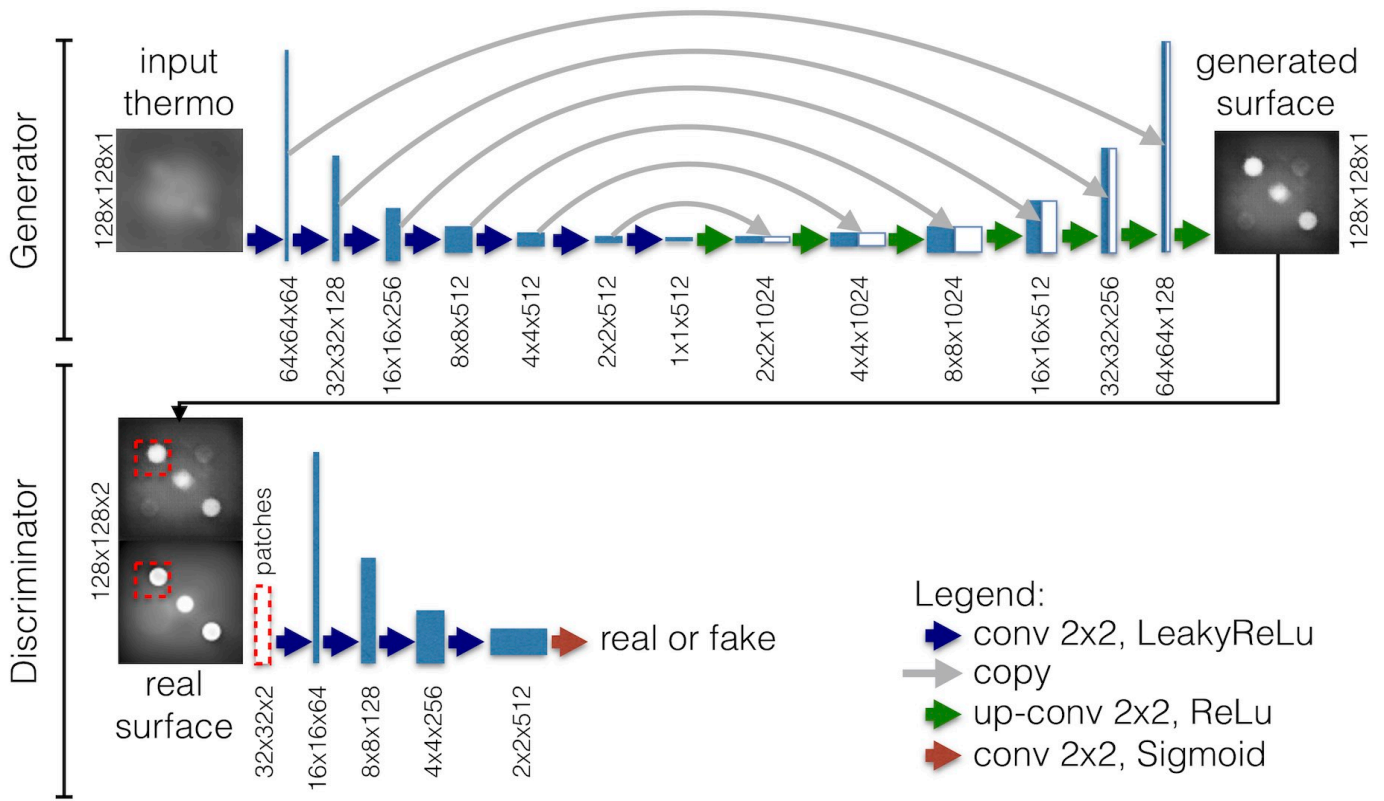

Fig. 2. Pix2pix UNet_128 GAN network architecture.

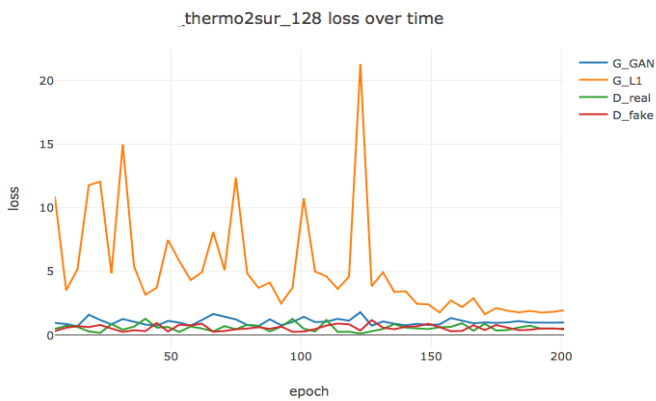

Fig. 3. Training loss over epoch.

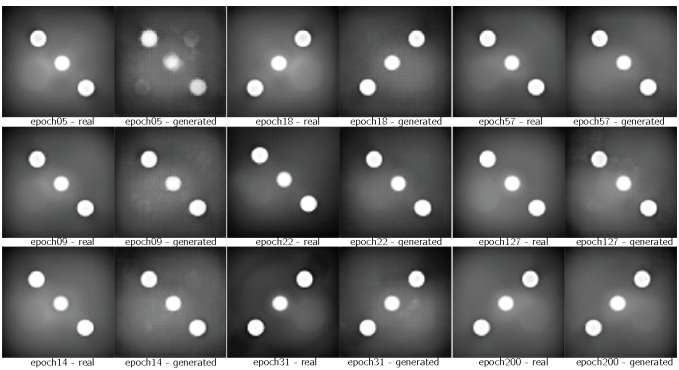

Fig. 4. Real/generated image comparison during the network training.

For our problem, the patch size will certainly change the prediction performance of various geometrical scale. The generator and the discriminator have the same layer architecture: each layer has a convolution layer and LeakyReLu [14] as an activation function. The augmentation part of the generator use Rectified Linear Unit [13]. In our study, we do not use any batch training nor normalization because we have a small dataset. Finally, pix2pix introduces a L1 distance loss between the generated images and the real ground truth images. Thus, the generator must fool the discriminator and minimize the L1 Manhattan distance with the real image. The complete network architecture is show in Fig. 2. We have only trained two hundred epochs for now. We use Xavier layers initialization [15] based on success in the literature. We envisage hyper parameters tuning as a direct pursuit of this research. Other architectures must also be evaluated, particularly the Perceptual Adversarial Network [16], which shows better performance in specific cases.

*D. Pix2pix GAN training*

The network was trained using the Adam stochastic gradient descent solver [17]. We kept the image augmentation techniques used in the pix2pix [7] proposition: random jitter and mirroring. The network training takes 3 hours on a Core i7 CPU. We will investigate multiple hyper parameters tuning as soon as we have new GPU computing resources. We observe the importance of the Conditional L1 loss in the training process (Fig. 3, Fig. 4). The training completion was effective after 180 epochs. The training is relatively fast. It can be done over a night for industrial applications.

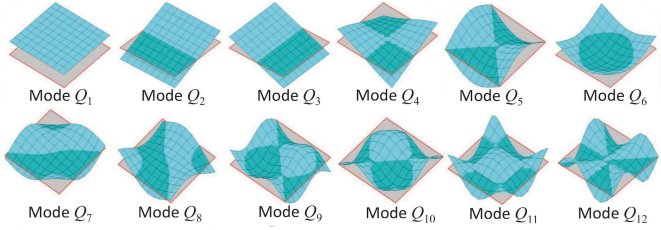

Fig. 5. Vector basis of modals descriptors (DMD) for a plane geometry [18].

*E. Discrete Modal Decomposition*

In order to describe geometrically the surfaces and check if the GAN generation enables to retrieve geometrical features from the true images, we introduce here some shape parameterization methods, especially the *Discrete Modal Decomposition* (DMD). Global shape parameterization techniques lay on the identification of parameters which characterize the shape geometrical elements as well as possible. In practice, we find descriptors by decomposing the surface into a descriptors space, specific to each method.

The DMD uses a decomposition space based on a vibration mechanics problem [19], [20]. Among the other widespread methods used for parameterization in the literature:

i. The *Discrete Cosine Transform* (DST) decomposition [21] uses a cosine harmonics space to describe the surface.
ii. The *Fourier series decomposition* is based on the Discrete Fourier Transform (DFT). This method enables to decompose a surface into a sine and cosine harmonics space, more general than the DST one.
iii. The *Spherical harmonics decomposition* describes complex shapes into a spherical harmonics space. This technique is especially used to characterize 3D shapes.

If we work with a plane geometry, the DMD descriptors are the natural vibration eigenmodes of the reference plane geometry (Fig 5). Le Goïc [18] has shown that this method can be used to characterize global as well as local surfaces geometries. We propose to use this method to parameterize the geometry of our samples, and to see if this parameterization can be retrieved through the GAN image generation.

IV. PREDICTION PERFORMANCE ANALYSIS

We use multiple methods to determine the predictive performance of the network. We firstly compare real measured height map images with images generated by the GAN (Fig. 6). The validation is done on a dedicated dataset of 14 parts (Fig 7). Simple statistical features and histogram comparisons with different metrics are used. Then, we introduce Discrete Modal parameters as an innovative prediction performance tool on specific geometrical features.

*A. Statistical and Haralick features comparisons*

Generated images visually show good similarities with real images. In order to validate the similarity, we compute different statistical features on each couple of real and generated images. We compute the mean, median, kurtosis, skewness and multiple quantile of the images. We also compute the Haralick features [22] as they are good textural descriptors. We study the difference between those values. We compute the p-value to validate the result on this really small population (14 elements). The standard deviation on all the tested results is high and the p-value was superior to 5% on most of the metrics. Thus, we cannot reject the null hypothesis [23]. We then must increase the test dataset size to be able to get significant results with these textural parameters.

The only features with a p-value superior to 5% are Haralick's difference variance, energy, entropy and homogeneity. The energy and the homogeneity metrics indicate the uniformity of the image. The entropy measured the presence of random patterns. Thus, our real and generated images are similar in their variance distribution.

*B. Real vs generated image similarity*

We avoided pixel to pixel comparison because our generated images can be slightly translated due to the dataset augmentation methods used. Histogram comparisons are robust to scaling and movements. We used various distance metrics on histograms (Fig. 8): Bhattacharya [24], Khisquared, correlation [25], cosine (similar to Euclidian L2), Hellinger, Kullback-Leibler [26], Manhattan L1 and Minkowski distances.

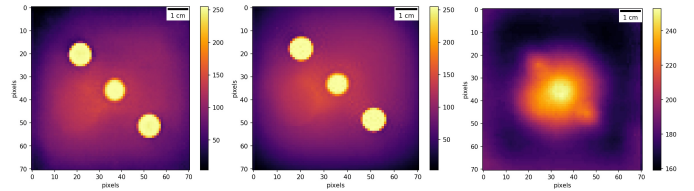

Fig. 6. Right: real geometry image, center: generated geometry image, left: inputed thermographic image (images are normalized between 0-255).

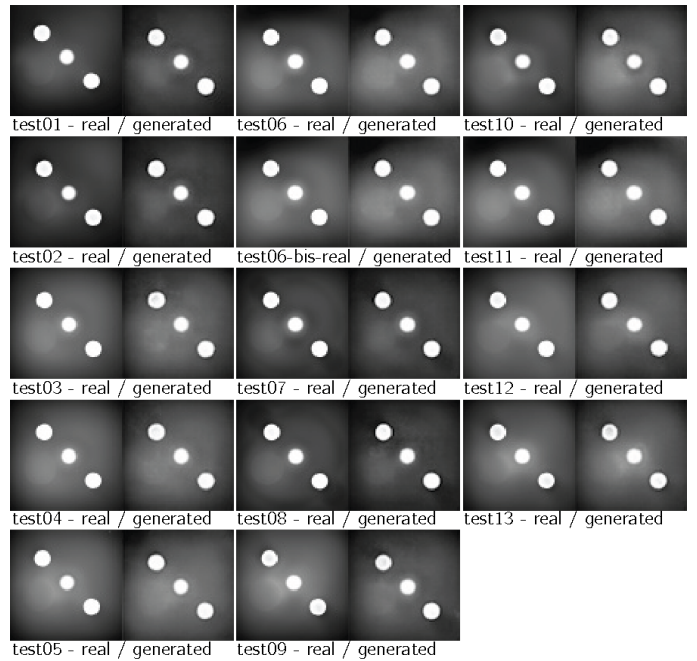

Fig. 7. Complete test dataset for the trained network after 200 epoch

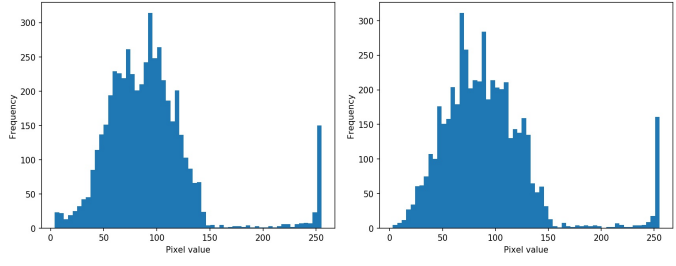

Fig. 8. Histogram comparison for test01: left is real, right is generated image

We used the p-value test to test for the null hypothesis and only the cosine and correlation distances were kept. We also compute the Structural Similarity [27] and the Peak Signal to Noise Ratio (PSNR), two widespread measures for image comparison. Results on the test dataset are shown in Table I. test06 and test06-bis are images obtained for the same settings in the design of experiment: no part with similar settings were present in the train dataset. Results on the test06 and test06-bis are used to study the generalization capability of this predictive method. Although test06 results are good, test06-bis is not on the same level of prediction quality based on cosine and correlation distances on histograms. The median value on the 14 parts shows the overall good predicting performance of the methods. Further analysis is done with the Discrete Modal Decompositions.

TABLE I.  IMAGES SIMILARITIES

| Part name | Similarity between real and generated images | | | |
|---|---|---|---|---|
| | Cosine distance | Correlation distance | PSNR | SSIM |
| test01 | 0.95 | 0.92 | 14.24 | 0.79 |
| test06 | 0.90 | 0.83 | 17.91 | 0.84 |
| test06-bis | 0.60 | 0.40 | 13.59 | 0.81 |
| MEDIAN 14 parts | 0.90 | 0.84 | 18.35 | 0.86 |
| STD 14 parts | 0.10 | 0.15 | 3.95 | 0.07 |

### C. Discrete Modal Decomposition analysis

We introduce here a new performance indicator for the GAN predictions: the DMD analysis. The objective is to study if the GAN is able to retrieve the surfaces in the DMD geometrical parameterization. DMD analysis is especially interesting in the case of plastic parts deformation. Our objective is to anticipate deformation on the real part. The firsts DMD coefficients give the global part geometry. The last DMD coefficients concern the roughness and very local deformation.

In plastic molding, we particularly have to anticipate the global part geometry, so, the first modal coefficients are especially appropriate.

The real pictures modal spectrums are calculated and compared with the generated images ones (Fig. 9). Similarity between the 2 spectrums vectors for the 14 pairs of images have been calculated thanks to correlation coefficient and cosine similarity (Table II). The results of the spectrums comparison study show a very good similarity between the spectrums with a median $R^2$ coefficient value of 0.98 and a median cosine distance value of 0.99 (very close to 1). Moreover, standard deviations are very low for both measurements, which shows a strong consistency for the 14 tested parts results. These measures bring to light the good performances of the GAN to retrieve the geometric parameterization of the parts' surfaces. It also shows the opportunity of using DMD analysis as an innovative performance indicator for neural network image generation. This paper illustrates a preliminary study for the use of GAN to predict geometry parameters through DMD. For further developments, we also propose to study these results from "the modes viewpoint". Instead of analyzing the correlation between images global modal spectrums, we want to investigate each mode independently to see which ones are best predicted by the network. Through this work, we hope to show what types of shapes and deformations the network is able to generate the best. As explained in III.E., DMD is not the only method to geometrically characterize surfaces. It could be also interesting to apply other type of decompositions and compare with the DMD analysis.

TABLE II.  SPECTRUMS SIMILARITY

| Part name | Similarity between real and generated images' spectrums vectors | |
|---|---|---|
| | Cosine distance | Correlation coefficient $R^2$ |
| test01 | 0.94 | 0.88 |
| test06 | 0.98 | 0.96 |
| test06-bis | 0.99 | 0.99 |
| MEDIAN 14 parts | 0.99 | 0.98 |
| STD 14 parts | 0.08 | 0.04 |

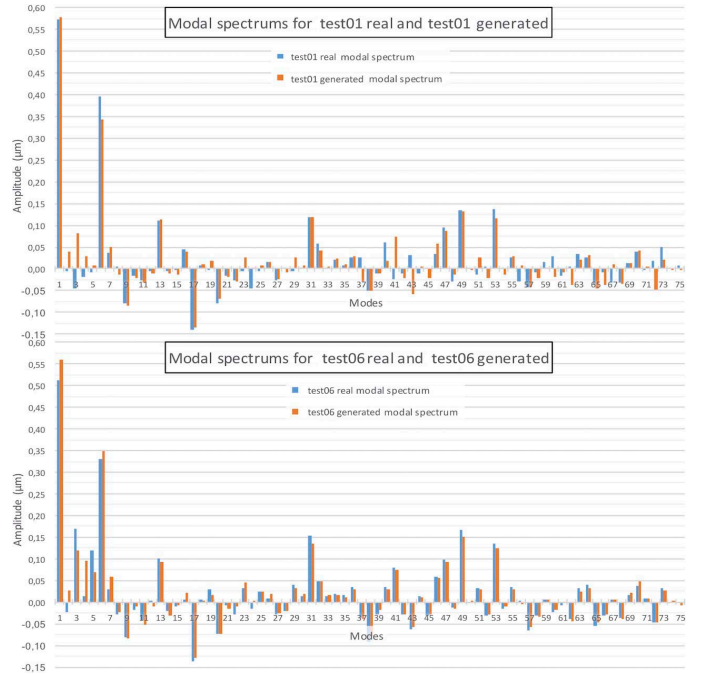

Fig. 9. DMD modal spectrum real/generated images comparison

## V. Conclusion

In this study, we have shown the possibility of using Generative Adversarial Networks pix2pix on a very small dataset, to comply with industrial constraints. Based on the measured similarity between final image and generate image, we show that the prediction is acceptable in the context of our very limited learning study. Quantities of information contained in the thermal image seem sufficient to predict our geometry in our specific application case. We introduce Discrete Modal Decomposition (DMD) to analyze the prediction performance from a geometrical parameterization side. The DMD presents a clear meaning to analyze geometrical properties but it could also further be used to analyze any generative network performance. In thermoplastic injection molding, we don't need to predict the rugosity (high modes in DMD) but only geometric form defects (low modes in DMD). Our problematic is to anticipate the global cold part geometry, so, the first modal coefficients are appropriate. Performance evaluation and hyperparameters tuning must be done on a larger dataset. Further studies can also be designed to analyze more precisely the GAN performances to retrieve independent modes to show what shapes can be best predicted. Finally, further research needs to be done on the complete 3D model generation based on thermographic prior, as literature shows encouraging result in 2D images to 3D shapes transformation [28].